\documentclass[sigconf,screen,nonacm]{acmart}

\usepackage{lmodern}
\usepackage{microtype}
\usepackage[T1]{fontenc}
\usepackage{booktabs}
\usepackage{multirow}
\usepackage{graphicx}
\usepackage{subcaption}
\usepackage{amsmath}
\usepackage{enumitem}
\usepackage{tikz}
\usetikzlibrary{shapes,arrows.meta,positioning,calc,fit,backgrounds}

\definecolor{emerald}{RGB}{16,185,129}
\definecolor{indigo}{RGB}{99,102,241}

\setlist[itemize]{nosep,leftmargin=*,topsep=1pt}
\setlist[enumerate]{nosep,leftmargin=*,topsep=1pt}

\AtBeginDocument{%
  \setlength{\abovedisplayskip}{3pt plus 1pt minus 1pt}%
  \setlength{\belowdisplayskip}{3pt plus 1pt minus 1pt}%
  \setlength{\abovedisplayshortskip}{2pt plus 1pt minus 1pt}%
  \setlength{\belowdisplayshortskip}{2pt plus 1pt minus 1pt}%
  \setlength{\bibsep}{0.5pt}%
}

\begin{document}

\title[WebOverlay: Vision-Only Web Overlay Detection]{WebOverlay: High-Throughput Vision-Only Distillation for Extreme Aspect-Ratio Web Overlay Detection}

\author{Jiannan Huang}
\email{jiannanh@google.com}
\affiliation{%
  \institution{Google}
  \country{USA}
}

\author{Jerry Ding}
\email{jerryd@google.com}
\affiliation{%
  \institution{Google}
  \country{USA}
}

\author{Li Yu}
\email{lyiu@google.com}
\affiliation{%
  \institution{Google}
  \country{USA}
}

\author{Zhenting Li}
\email{zhenting@google.com}
\affiliation{%
  \institution{Google}
  \country{USA}
}

\settopmatter{printccs=false,printacmref=false}
\renewcommand\footnotetextcopyrightpermission[1]{}
\acmConference[WWW '27]{Proceedings of the ACM Web Conference 2027}{2027}{}

\begin{abstract}
Overlays such as cookie-consent banners, modal dialogs, and translucent backdrops hide the content that web-scale pipelines crawl and index. Detecting them in full-page screenshots is hard because pages reach extreme aspect ratios: squashing a page into a square input erases small dismiss controls, while uniform tiling makes cost grow with page length. Multimodal LLMs (MLLMs) are accurate but too slow and costly at crawl scale, and HTML rules and lightweight CNNs miss many overlays. We distill a commercial MLLM teacher into a vision-only SigLIP student that scores full-width crops at the top, center, and bottom of the page, where viewport-pinned overlays land once the renderer expands the viewport to the full document. Cost remains constant for any page length, and because the crops are scaled by page width alone, at inference the student sees long pages at higher resolution than its teacher, which receives a downscaled whole page. On our human-verified web overlay benchmark, the student closely approaches its teacher in overall accuracy while maintaining high recall on challenging long pages. We also find that pre-filling annotators with teacher verdicts introduces measurable anchoring bias toward the teacher. Deployed on Cloud TPU accelerators, the student operates at a small fraction of its teacher's serving cost.
\end{abstract}


\maketitle

\begin{figure*}[t]
\centering
\includegraphics[width=0.98\textwidth]{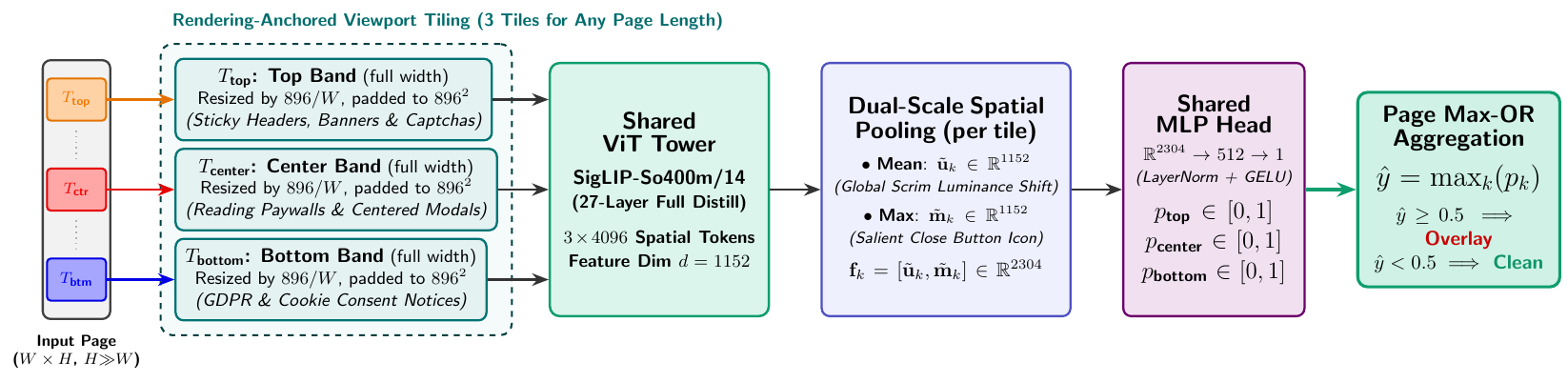}
\vspace{-4pt}
\caption{\textbf{Framework overview.} Rendering-anchored viewport tiling cuts three full-width bands ($T_{\text{top}}, T_{\text{center}}, T_{\text{bottom}}$) from an arbitrarily long page (e.g., $\rho = H/W = 15$); each band is resized by $896/W$ and padded to $896^2$, so cost is three encoder passes for any page length. Each tile is processed by a shared, fully fine-tuned 27-layer SigLIP ViT with dual-scale spatial pooling ($\mathbb{R}^{2304}$) and a shared MLP head, and tile scores are combined by page-level Max-OR aggregation.}
\Description{Overview diagram of rendering-anchored viewport tiling feeding a shared SigLIP encoder, dual-scale pooling, a shared MLP head, and Max-OR aggregation.}
\label{fig:teaser_arch}
\vspace{-4pt}
\end{figure*}

\section{Introduction}
\label{sec:intro}

Websites today are filled with visual interruptions. Driven by privacy statutes like GDPR and CCPA, as well as monetization gates, modern sites routinely inject modal banners, full-screen dark scrims, cookie prompts, and anti-bot verification challenges. While intended for human visitors, these floating elements severely disrupt automated web-scale systems. Search engine indexers, accessibility tools, and vision-language agents all suffer: an undetected overlay occludes main-body content and corrupts visual embeddings.

Detecting overlays in high-throughput crawler streams exposes an acute trilemma across \textit{accuracy}, \textit{latency}, and \textit{economic cost}:
\begin{enumerate}[leftmargin=*]
    \item \textbf{Multimodal LLMs}: Vision-language models (e.g., GPT-4o, Gemini 2.5 Flash) provide strong visual reasoning, but querying commercial APIs at crawl scale is prohibitive: each request costs hundreds of milliseconds of API roundtrip (\S\ref{subsec:cost}), and billing grows linearly with a crawl that indexes billions of URLs.
    \item \textbf{HTML DOM Heuristics}: Rules over the rendered HTML and its CSS (\texttt{position: fixed}, \texttt{z-index}) are fast but hit an information ceiling: they miss a third of overlays while flagging many benign floating elements. Part of this ceiling is structural: consent managers and anti-bot challenges often render inside cross-origin \texttt{<iframe>} elements or closed shadow roots, which leave no accessible nodes in the host DOM.
    \item \textbf{Lightweight Vision Classifiers}: Small convolutional networks (e.g., MobileNetV4~\cite{qin2024mobilenetv4}) are equally fast but collapse when long pages are compressed into a single square input.
\end{enumerate}

The root cause of this vision failure is page geometry. Web screenshots are not natural square images: page aspect ratios in our corpus reach $20{:}1$, and a crawler cannot know page length in advance. Downsampling a long capture to a square input shrinks its width to a narrow strip, blurring text, dialog borders, and dismiss buttons beyond recognition. Conversely, uniform grid tiling (e.g., LLaVA-1.5-HD~\cite{liu2024improved}) makes the token count grow with page length---covering the longest page in our corpus with viewport-sized tiles takes 27 tiles---and can split a dialog across tiles.

We address this with a vision-only student distilled from an MLLM teacher. Our renderer expands the viewport to the full document height, so overlays pinned with \texttt{position: fixed} or \texttt{sticky} land at the top, center, or bottom of the page. The student therefore scores three full-width crops at these positions. Its cost is constant, and because the crops are scaled by page width alone, the student sees long pages at higher resolution than its teacher, which receives a downscaled whole page. Our contributions are:
\begin{enumerate}[leftmargin=*]
    \item \textbf{Rendering-anchored constant-cost tiling.} We derive three anchor positions from full-document frame expansion and score one full-width crop at each. Cost stays at three encoder passes for any page length, whereas covering a page with viewport-sized tiles takes 4 tiles for the median page in our corpus and up to 27. The three crops recall $86.9\%$ of overlays, versus $71.7\%$ for one whole-page view of the same model.
    \item \textbf{Resolution asymmetry.} The teacher reads a downscaled page; at inference the student reads crops whose scale does not shrink with page length. As aspect ratio grows, teacher recall falls from $99.4\%$ to $59.1\%$, the same student given one whole-page view falls from $80.7\%$ to $36.4\%$, and the three-crop student stays above $81\%$. On an independent blind benchmark, the student matches the teacher on standard pages and decisively outperforms it on long pages ($p < 0.001$, $+55.7$ F1 points).
    \item \textbf{Anchoring bias in LLM-assisted labels.} On a paired $289$-page subset re-annotated blindly without model hints, pre-filling annotators with the teacher's verdict inflates teacher agreement on long pages ($\kappa$ $0.59 \to 0.88$) and reverses the student--teacher ordering ($92.9\%$ vs.\ $63.6\%$ F1 under blind labels), as confirmed on an independent $592$-page blind benchmark (\S\ref{subsec:blind}).
    \item \textbf{Serving economics and a negative result.} On Cloud TPU v5e the student costs $\$63$ per million pages, 83\% below its teacher's list price and about 80\% below the cheapest current Gemini model (\S\ref{sec:deployment}). Cheap pre-filters rank overlays poorly, so a two-tier cascade saves traffic only by discarding overlays; we recommend single-tier serving for audit-grade crawling.
\end{enumerate}

\section{Related Work}
\label{sec:related_work}

\subsection{Web Content Extraction and Consent Auditing}
\label{subsec:rw_web_extraction}
Extracting the main content of a page from raw HTML has a long history in information retrieval. Early approaches used hand-crafted rules, tag density, and text-to-tag ratios~\cite{gottron2008content,gupta2003dom} to discard navigation bars, footers, and boilerplate; visual page segmentation (e.g., VIPS~\cite{cai2003vips}) partitions rendered DOM structures into blocks using CSS geometry. More recent work embeds DOM trees with transformer encoders (DOM-LM~\cite{deng2022dom}, MarkupLM~\cite{li2022markuplm}) for node classification, and systems such as CookieEnforcer~\cite{khandelwal2023cookieenforcer} and AdHere~\cite{yan2023adhere} detect cookie notices or intrusive interstitials from rendered DOM trees, mutation observers, and interaction modeling. All of these depend on an accessible DOM, whereas consent-management platforms and overlay scripts often use randomized class names, closed shadow roots, and cross-origin iframes. On our evaluation set, HTML rules reach only $44.4\%$ F1 (\S\ref{subsec:main_results}), which motivates working from rendered pixels.

\subsection{UI Element Detection and Visual Pop-up Blocking}
\label{subsec:rw_ui_detection}
Datasets such as Rico~\cite{deka2017rico} and WebUI~\cite{wu2023webui} support detecting buttons, icons, and text fields on screen captures, typically by adapting object detectors such as Faster R-CNN~\cite{ren2015faster} or the YOLO series~\cite{jocher2023ultralytics}, and Screen2Vec~\cite{li2021screen2vec} embeds whole GUI screens. Perceptual ad blockers such as Percival~\cite{din2020percival} and mobile GUI tools such as PopSweeper~\cite{guo2024popsweeper} use computer vision to detect ads and app-blocking pop-ups. These tools target isolated elements or single-viewport screens. Full-page web captures add two difficulties: overlays often include semi-transparent scrims that span the whole document and have no tight bounding box, and squashing a page with an aspect ratio up to $20{:}1$ into a square input shrinks dismiss controls below recognition.

\subsection{Vision-Language Models and Lightweight Distillation}
\label{subsec:rw_distillation}
Vision-language encoders (CLIP~\cite{radford2021learning}, SigLIP~\cite{zhai2023sigmoid}) and multimodal LLMs (PaliGemma~\cite{beyer2024paligemma}, LLaVA~\cite{liu2024visual}, Gemini~\cite{geminiteam2024gemini}) generalize well across visual question answering and scene understanding. General-purpose models handle variable aspect ratios by patch packing (NaViT~\cite{dehghani2023patch}) or uniform grid tiling (e.g., LLaVA-1.5-HD~\cite{liu2024improved}), so their token count grows with page length. Knowledge distillation~\cite{hinton2015distilling,gou2021knowledge} compresses a heavy teacher into a compact student and has been studied for image classification (DeiT~\cite{touvron2021training}) and vision-language models (DistillVLM~\cite{fang2021compressing}); distilling overlay detection from a generative MLLM into a ViT encoder has, to our knowledge, not been studied.

Students can outperform their teachers. In Born-Again Networks~\cite{furlanello2018born} and Noisy Student~\cite{xie2020self}, the gain comes from retraining on the teacher's outputs, with added noise and unlabeled data in the latter, and in weak-to-strong generalization~\cite{burns2024weak} from a more capable pretrained student. Learning using privileged information~\cite{lopezpaz2016unifying} studies an input asymmetry in the other direction: the teacher sees richer input than the student. Our setting reverses this asymmetry. The teacher sees a downscaled page, whereas at inference the student sees width-aligned crops, which are sharper than the teacher's input on all but the shortest pages. The student approaches the teacher's accuracy at a small fraction of its serving cost (\S\ref{sec:deployment}) and loses far less recall on long pages (\S\ref{subsec:slices}).

\section{Methodology}
\label{sec:methodology}

We deliberately keep the neural components of the student standard: mean/max token pooling and max aggregation are common multiple-instance building blocks. Our core methodological contribution is the rendering-derived choice of where to look, and the resolution asymmetry it creates between student and teacher.

\subsection{Problem Formulation}
\label{subsec:formulation}
Let an arbitrary rendered webpage screenshot be denoted as a continuous RGB tensor $\mathbf{I} \in \mathbb{R}^{H \times W \times 3}$, where $W$ is the rendered viewport width and $H$ is the full scroll height after full-document frame expansion. We define the physical aspect ratio as $\rho = H / W$. Across WebOverlay-2.4K, rendered widths range from $1024$ to $3840\,\text{px}$: $96\%$ of pages ($2{,}263$/$2{,}364$) lie between $1024$ and $1920\,\text{px}$, $1920\,\text{px}$ is the most common width ($945$ pages, $40\%$), and $101$ pages ($4.3\%$) are $2\times$ high-DPI captures ($2048$--$3840\,\text{px}$). This heterogeneity makes fixed-pixel cropping fragile and motivates width-aligned preprocessing.

Aspect ratios follow a pronounced long-tail distribution: the median page is $\rho = 2.79$, but $11.6\%$ of pages exceed $\rho = 8$ and the longest reaches $\rho = 20.6$. Because crawl pipelines must apply a uniform ingestion policy, tuning for the median degrades the long tail. We therefore stratify the evaluation set into \textit{Test-Common} ($N{=}1{,}873$) and \textit{Test-Hard} ($N{=}491$, enriched for elongated pages) to evaluate both regimes.

Our objective is to learn a mapping function $f_\theta: \mathbf{I} \mapsto \hat{y}_{\text{page}} = P(\text{has\_overlay} = 1 \mid \mathbf{I}; \theta) \in [0, 1]$, where $\hat{y}_{\text{page}}$ denotes the predicted probability that the webpage contains an intrusive blocking overlay (any floating, sticky, or fixed-position UI component that obstructs the primary document body, including cookie consent notices, GDPR banners, authentication/newsletter modals, and anti-bot CAPTCHA challenges).

\subsection{Rendering-Anchored Viewport Tiling}
\label{subsec:tiling}
The encoder takes a fixed $S \times S$ input with $S = 896$. Fitting a whole page into it scales the page by $S / \max(W, H)$: on a $1920$\,px-wide page with $\rho = 8$, a $30$\,px dismiss button shrinks to under $2$\,px. Covering the page with viewport-sized tiles avoids this, but cost then grows with page length; with the viewport height $h$ defined below, full coverage takes a median of $4$ tiles per page in WebOverlay-2.4K and up to $27$.

We instead score three crops at fixed positions. Let $h = \min\left(H, \max(0.75\,W, 800)\right)$ be the height of one landscape viewport ($h = 0.75\,W$ for $95\%$ of pages). We cut three full-width bands of height $h$, indexing rows of $\mathbf{I}$: $B_{\text{top}} = \mathbf{I}[0 : h]$, $B_{\text{center}} = \mathbf{I}[(H{-}h)/2 : (H{+}h)/2]$, and $B_{\text{bottom}} = \mathbf{I}[H{-}h : H]$. Each band is resized with its aspect ratio preserved and centered on a gray $S \times S$ canvas, giving tiles $T_{\text{top}}, T_{\text{center}}, T_{\text{bottom}}$. Because $h \le W$, the resize factor is $S / W$ (e.g., $0.47$ for a $1920$\,px page), and the content occupies $896 \times 672$ pixels when $h = 0.75\,W$. The factor depends on page width only, so an element keeps the same size in its tile however long the page is. Tiles are therefore $\max(1, \rho)$ times sharper than the whole-page view, and $0.7 \max(1, \rho)$ times sharper than the teacher's input, whose longer side is resized to $1{,}280$\,px (\S\ref{subsec:setup}); the tiles are sharper than the teacher's view on $83\%$ of pages ($\rho > 1.43$).

The positions follow from how pages are rendered. The renderer expands the viewport to the full document height before capture, so an element styled \texttt{position: fixed; bottom: 0} lands at the bottom of the screenshot, one styled \texttt{top: 0} at the top, and a dialog centered in the viewport lands at the vertical middle; a full-viewport scrim covers the whole page. The three bands target these positions, and cost stays at three encoder passes for any page length.

\noindent\textbf{Why these three positions?} For $k = 3$, top, center, and bottom coincide with an evenly spaced placement, so the geometry itself is not new. The rendering argument adds that these positions stay sufficient however long the page is, which is what keeps cost constant. Tile activations are consistent with this: of the $420$ overlay pages, the bottom tile fires (score $\ge \tau$) on $69.0\%$, the center tile on $22.1\%$, and the top tile on $12.9\%$, and $57.1\%$ are caught by the bottom tile alone. Activations show where the model fires rather than where overlays are; measuring overlay positions directly would require layout boxes, which our DOM snapshots do not contain. Simpler alternatives have clear gaps. The first three viewports never reach the page bottom on $1{,}431$ of the $2{,}364$ pages, which contain $228$ of the $420$ overlay pages. Random crops are not reproducible across runs. Full coverage needs $1.9\times$ as many encoder passes on average and up to $9\times$ on the longest pages; we have not measured whether it recovers the overlays our three tiles miss.

\subsection{Dual-Scale Spatial Pooling}
\label{subsec:pooling}
Each tile $T_k \in \{T_{\text{top}}, T_{\text{center}}, T_{\text{bottom}}\}$ is independently processed by a shared 27-layer Vision Transformer backbone (SigLIP-So400m/14, patch size $P=14$), outputting spatial token feature matrices:
\begin{equation}
\mathbf{Z}_k = \mathrm{ViT}(T_k) \in \mathbb{R}^{N \times d}, \; N = (896/14)^2 = 4096, \; d = 1152
\end{equation}
Overlay evidence appears at two scales: a semi-transparent backdrop that darkens much of the tile, spread over thousands of tokens, and small high-contrast controls, such as a dismiss button under $30$\,px, that cover only a few tokens. We therefore summarize each tile with two standard poolings (\emph{dual-scale pooling}): (1) \textbf{Mean pooling} $\tilde{\mathbf{u}}_k = \bar{\mathbf{z}}_k / (\|\bar{\mathbf{z}}_k\|_2 + \epsilon)$ with $\bar{\mathbf{z}}_k = \frac{1}{N} \sum_{i=1}^N \mathbf{Z}_{k, i}$, which responds to broad changes such as backdrop darkening; and (2) element-wise \textbf{Max pooling} $\tilde{\mathbf{m}}_k = \hat{\mathbf{z}}_k / (\|\hat{\mathbf{z}}_k\|_2 + \epsilon)$ with $\hat{\mathbf{z}}_k = \max_{i=1 \dots N} \mathbf{Z}_{k, i}$, which keeps strong local activations from small controls that mean pooling would attenuate by up to a factor of $N = 4096$. We concatenate the two into $\mathbf{f}_k = [\tilde{\mathbf{u}}_k, \; \tilde{\mathbf{m}}_k]^\top \in \mathbb{R}^{2304}$, and a small MLP head with LayerNorm and GELU outputs the tile probability:
\begin{equation}
p_k = \sigma\left(\mathbf{W}_2 \text{GELU}(\text{LN}(\mathbf{W}_1 \mathbf{f}_k + \mathbf{b}_1)) + b_2\right) \in [0, 1]
\end{equation}
where $\mathbf{W}_1 \in \mathbb{R}^{512 \times 2304}$ and $\mathbf{W}_2 \in \mathbb{R}^{1 \times 512}$. Using both poolings is a design choice motivated by the two scales above; we have not ablated either pooling on its own.

\subsection{Page-Level Max-OR Aggregation}
\label{subsec:aggregation}
An overlay in any one tile is enough to obstruct the page, so we score the page by the maximum tile probability (Max-OR), the standard max rule of multiple-instance learning: $\hat{y}_{\text{page}} = \max(p_{\text{top}}, p_{\text{center}}, p_{\text{bottom}})$, classifying a page as obstructed if $\hat{y}_{\text{page}} \ge \tau$. The rule needs no tile weighting, keeps tiles independent so they can be batched in parallel, and attributes each positive decision to a specific tile.

\subsection{End-to-End Distillation Objective}
\label{subsec:distillation}
Pre-trained on natural images, SigLIP is not natively sensitive to rendered UI artifacts such as translucent scrims; in \S\ref{subsec:ablations}, adapting only the top four layers lowers F1 by $2.9$ points at the fixed threshold, although AP changes little. We therefore unfreeze all 27 transformer layers of SigLIP and train end-to-end on the teacher's verdicts, i.e., hard-label distillation~\cite{touvron2021training}.

The teacher is Gemini 2.5 Flash~\cite{comanici2025gemini25}, prompted with a fixed overlay rubric. For each training page it returns a binary verdict $y \in \{0, 1\}$ and a score $p_{\text{teacher}}$; in our training set the score takes only two values, $0.95$ for positive and $0.05$ for negative verdicts. The teacher's verdicts serve as the training labels. During training, each page is shown to the student as a single whole-page view, downscaled to fit the $S \times S$ input, and paired with these page-level targets; no tile-level labels are used. Tiling and max aggregation are applied only at inference, where the same network scores each tile. The student is thus trained at a lower resolution than it is tested at, a train--test discrepancy of the kind studied by FixRes~\cite{touvron2019fixing}; unlike FixRes, we do not fine-tune at the test resolution. Writing $z$ for the student logit and $\ell(z, t) = -t \log \sigma(z) - (1 - t) \log(1 - \sigma(z))$ for binary cross-entropy, the objective combines a class-weighted hard-label term with a term on the teacher's score:
\begin{equation}
\mathcal{L}(\theta) = 0.70 \cdot \ell_{w}(z, y) + 0.30 \cdot \ell(z, p_{\text{teacher}})
\end{equation}
where $\ell_{w}$ up-weights the positive term by $w_{\text{pos}} = 2.5$ to counteract class imbalance. Because $p_{\text{teacher}}$ is a fixed function of $y$, the second term acts as label smoothing rather than as distillation from graded teacher confidences. We optimize with AdamW (weight decay $0.05$) under a cosine learning-rate schedule. To fit full 27-layer backpropagation at $896^2$ resolution into accelerator memory, we recompute block activations during the backward pass (activation rematerialization).

\begin{table*}[t]
\centering
\small
\caption{Main comparison on the human-verified \textbf{WebOverlay-2.4K} evaluation set (2,364 pages; 420 overlay, 1,944 clean; teacher-assisted labels). All metrics in \%. All score-producing models use the same fixed operating point $\tau = 0.5$; no threshold is tuned on the evaluation set. F1 CI: percentile bootstrap $95\%$ interval over pages ($10{,}000$ resamples).}
\label{tab:main_results}
\setlength{\tabcolsep}{4pt}
\begin{tabular}{llcccccc}
\toprule
\textbf{Paradigm} & \textbf{Model} & \textbf{F1} & \textbf{F1 95\% CI} & \textbf{Precision} & \textbf{Recall} & \textbf{Acc} & \textbf{ROC-AUC} \\
\midrule
\textit{HTML Rules} & Regex Rules over Rendered HTML\textsuperscript{\dag} & 44.44 & [41.0, 47.8] & 33.33 & 66.67 & 70.39 & --- \\
\textit{Lightweight CNN} & MobileNetV4 & 31.47 & [28.9, 34.0] & 19.49 & 81.67 & 36.80 & 62.60 \\
\textit{Object Detector} & YOLO-style ConvNet Detector\textsuperscript{\S} & 39.50 & [36.0, 42.9] & 29.38 & 60.24 & 67.22 & 69.75 \\
\textit{Commercial MLLM} & Gemini 2.5 Flash (Teacher)\textsuperscript{\ddag} & \textbf{89.74} & [87.6, 91.8] & 85.22 & \textbf{94.76} & \textbf{96.15} & --- \\
\midrule
\textit{Ours} & SigLIP-27L Distilled, Whole-Page View & 81.57 & [78.4, 84.6] & \textbf{94.65} & 71.67 & 94.25 & 91.32 \\
\textbf{Ours} & \textbf{SigLIP-27L Distilled, 3-Tile Max-OR} & \textbf{85.98} & [83.4, 88.4] & 85.08 & 86.90 & 94.97 & \textbf{94.95} \\
\bottomrule
\end{tabular}

\vspace{2pt}
\parbox{0.95\textwidth}{\footnotesize $^\dagger$Regular expressions over the rendered HTML of every page (inline and embedded CSS only; no computed styles or layout boxes); strictest of three rule tiers (\S\ref{subsec:setup}); emits hard verdicts. $^\S$A four-stage convolutional backbone with a dense anchor-based box head trained in-house; page score = max box confidence. $^\ddagger$Our distillation teacher, which also produced the training labels; it emits binary verdicts, so ROC-AUC is not defined.}
\end{table*}

\section{Empirical Evaluation}
\label{sec:experiments}

\subsection{Experimental Setup}
\label{subsec:setup}
\noindent\textbf{Training Data}: The student is trained on $10{,}000$ examples covering $7{,}500$ unique rendered pages (some pages are repeated), labeled by the teacher MLLM (\S\ref{subsec:distillation}). Training and evaluation pages were sampled from different shard ranges of the same production render store. A post-hoc check nevertheless found that $136$ of the $2{,}500$ sampled evaluation pages ($5.4\%$; $127$ from Test-Common, $9$ from Test-Hard) share a URL with a training page, where they carry the teacher's verdict as label. We remove these pages from the evaluation set; all results below use the remaining $2{,}364$ pages (\S\ref{subsec:main_results} reports the effect of keeping them).

\noindent\textbf{WebOverlay-2.4K Evaluation Set \& Rendering Protocol}: Existing document and UI datasets predominantly focus on mobile screens or synthetic web templates. We therefore assemble \textbf{WebOverlay-2.4K}, an internal evaluation set comprising $2{,}364$ full-page screenshots spanning $2{,}353$ domains sampled from a separate shard range of a production web crawl, of which $17.8\%$ contain an overlay. To avoid missing delayed JavaScript popups, pages are rendered by our production headless browser with JavaScript and cookies enabled, full-document frame expansion, and a generous render timeout that waits for network and layout quiescence. WebOverlay-2.4K consists of third-party page screenshots from an internal crawl and is not released; a public benchmark built from independently sampled pages is in preparation.

\noindent\textbf{Human Reference Labels \& Blind Benchmarks}: Every page in \textbf{WebOverlay-2.4K} was reviewed by a human annotator ($420$ overlay, $1{,}944$ clean). For efficiency, pages were presented with the teacher's verdict pre-filled, which the annotator confirmed or corrected on the full screenshot. Because pre-filling can anchor annotators toward the teacher, we conduct two teacher-free blind evaluations (\S\ref{subsec:blind}): (1) a paired $289$-page subset of WebOverlay-2.4K ($191$ Test-Common, $98$ Test-Hard) re-annotated blindly to isolate anchoring on identical pages, and (2) \textbf{\textsc{WebOverlay-OOD}} ($N{=}592$; $475$ Val-Common, $117$ Val-Hard), an independent blind validation and evaluation benchmark from a disjoint crawl partition ($\le 1$ page/domain; strictly disjoint in URL, SHA-256, and domain from training and WebOverlay-2.4K; $8$ broken/unsafe renders excluded from $600$ candidates). Both blind sets were labeled from raw screenshots alone without model scores, prior labels, or metadata.

\noindent\textbf{Teacher Validity \& Anchoring Bias}: Against teacher-assisted labels on WebOverlay-2.4K, the teacher agrees on $96.2\%$ of pages ($\kappa = 0.874$; $398$ TP, $69$ FP, $22$ FN, $1{,}875$ TN). On the paired $289$-page blind subset, however, $11$ of $12$ disagreements between assisted and blind labels sided with the pre-filled teacher verdict ($6/6$ on Test-Hard), inflating apparent teacher agreement on long pages from $\kappa = 0.59$ to $\kappa = 0.88$ ($63.6\% \to 88.9\%$ F1) and reversing the student--teacher ordering ($92.9\%$ vs.\ $63.6\%$ F1 under blind labels; McNemar $6$--$0$, $p = 0.031$). On \textsc{WebOverlay-OOD} (\S\ref{subsec:blind}), teacher accuracy on long pages drops further ($30.8\%$ F1, Recall $21.1\%$, $\kappa = 0.24$), while the student remains robust ($86.5\%$ F1, $\kappa = 0.84$). Anchoring on pre-annotations is well documented in linguistic annotation~\cite{fort2010influence,berzak2016anchoring}.

\noindent\textbf{Slices}: \textit{Test-Hard} ($N{=}491$) and \textit{Val-Hard} ($N{=}117$) collect pages with $\rho > 8$ or rendered height above $10{,}000$\,px (plus pages from selected non-English EU/JP ccTLDs); the remaining pages form \textit{Test-Common} ($N{=}1{,}873$) and \textit{Val-Common} ($N{=}475$).

\noindent\textbf{Baseline Models}: We compare against four representative paradigms:
\begin{enumerate}[leftmargin=*]
    \item \textit{HTML Rules}: hand-written regular expressions applied to the rendered HTML of every page. A page is flagged if it has a dialog or backdrop marker (an ARIA dialog role, a modal backdrop class, or a known consent-manager id) and a dismiss or accept control, or a visible modal-like container, a fixed-position CSS rule with a z-index of at least $100$, and such a control. Only inline and embedded CSS is visible to the rules; computed styles and layout boxes are not available. We report the strictest of three rule tiers; the two looser tiers reach $97$--$98\%$ recall at $20$--$22\%$ precision.
    \item \textit{Lightweight CNN}: MobileNetV4-Small~\cite{qin2024mobilenetv4} ($384\times 384$ input), a production legacy visual baseline previously deployed in web indexing pipelines.
    \item \textit{Object Detector}: a \textbf{YOLO-style Multi-Stage ConvNet Detector}~\cite{jocher2023ultralytics,ren2015faster} ($640\times 640$ input), trained in-house for bounding-box overlay localization. Page-level scores are obtained by taking the maximum box confidence over the page.
    \item \textit{Commercial MLLM}: Gemini 2.5 Flash, prompted with our popup-detection rubric; this is also our distillation teacher. It receives the whole screenshot with its longer side resized to $1{,}280$\,px, so a $1920$\,px-wide page with $\rho = 8$ reaches it only $160$\,px wide. It returns binary verdicts.
\end{enumerate}
Both vision baselines are previously deployed operational models rather than models tuned for this benchmark. Both take a single square input ($384^2$ and $640^2$), which shrinks small controls on long pages, and a box detector has no compact box to place around a full-page translucent scrim.

\noindent\textbf{Operating Point, Threshold-Free Evaluation \& Uncertainty}: Architecture, loss weights, epoch count, and the fixed operating point $\tau = 0.5$ were set prior to evaluation and validated on \textsc{WebOverlay-OOD} without tuning on WebOverlay-2.4K or its slices. Because single-threshold $F_1$ is sensitive to calibration, we additionally report threshold-free AP and ROC-AUC in Table~\ref{tab:threshold_free}. Confidence intervals are percentile intervals from $10{,}000$ bootstrap resamples of pages; differences between two models use paired resamples. MobileNetV4 is severely \textbf{mis-calibrated}, with all page scores collapsing into a narrow band around $0.5$, so its $F_1$ reflects where the cut lands inside this band rather than model quality; its ranking signal is weak but genuine (AUC $62.6\%$).

\begin{table}[t]
\centering
\caption{Threshold-free evaluation of the score-producing models on WebOverlay-2.4K (base rate in slice header). AP and AUC are computed from raw per-page scores and require no operating point; AP is given with its bootstrap $95\%$ CI. The HTML rules and the MLLM emit binary verdicts and are omitted.}
\label{tab:threshold_free}
\resizebox{\columnwidth}{!}{%
\begin{tabular}{llccc}
\toprule
\textbf{Evaluation Slice} & \textbf{Model} & \textbf{AP (\%)} & \textbf{AP 95\% CI} & \textbf{ROC-AUC (\%)} \\
\midrule
\multirow{3}{*}{\textit{Full-2.4K} ($17.8\%$)}
 & MobileNetV4 & 39.70 & [35.2, 44.3] & 62.60 \\
 & YOLO ConvNet & 40.46 & [35.6, 45.7] & 69.75 \\
 & \textbf{Ours (3-Tile)} & \textbf{87.78} & [84.5, 90.7] & \textbf{94.95} \\
\midrule
\multirow{3}{*}{\textit{Test-Common} ($20.1\%$)}
 & MobileNetV4 & 40.40 & [35.7, 45.0] & 60.87 \\
 & YOLO ConvNet & 41.86 & [36.7, 47.3] & 68.14 \\
 & \textbf{Ours (3-Tile)} & \textbf{90.52} & [87.4, 93.2] & \textbf{95.18} \\
\midrule
\multirow{3}{*}{\textit{Test-Hard} ($8.8\%$)}
 & MobileNetV4 & 41.65 & [27.2, 57.2] & 72.55 \\
 & YOLO ConvNet & 36.34 & [23.6, 53.5] & 74.61 \\
 & \textbf{Ours (3-Tile)} & \textbf{75.95} & [62.8, 87.1] & \textbf{96.93} \\
\bottomrule
\end{tabular}%
}
\end{table}

\subsection{Main Comparison}
\label{subsec:main_results}
Table~\ref{tab:main_results} presents the primary results on WebOverlay-2.4K.

\noindent\textbf{Limitations of HTML Rules \& Lightweight CNNs}: HTML rules cannot see content inside cross-origin iframes or closed shadow roots and fire on many benign fixed-position elements, MobileNetV4 loses small UI elements to downsampling blur on long pages, and the YOLO-style detector struggles because non-local translucent scrims lack tight bounding boxes. None exceeds $45\%$ F1.

\noindent\textbf{Approaching the Commercial MLLM}: Our 3-tile model reaches \textbf{86.0\% F1} and \textbf{94.9\% ROC-AUC}, $3.8$ points below the teacher ($89.7\%$ F1) at a fraction of its serving cost (\S\ref{subsec:cost}). Against the teacher-assisted labels, the teacher's lead on the full set is significant (paired $95\%$ CI of the F1 gap: $[-6.7, -0.9]$ points).

\noindent\textbf{Train--Evaluation Overlap}: Keeping the $136$ removed pages (\S\ref{subsec:setup}) moves our F1 from $86.0\%$ to $86.4\%$ and the teacher's from $89.7\%$ to $89.8\%$ (gap $-3.4$, CI $[-6.1, -0.7]$). Every comparison in \S\ref{sec:experiments} keeps its sign. The only change in significance is on Test-Common, where the teacher's lead shrinks to $2.4$ points and is no longer significant (CI $[-5.1, +0.2]$). On the paired in-set blind Test-Hard subset ($N{=}98$), keeping its $2$ overlapping pages adds one more discordant page, which the student wins ($7$--$0$, $p = 0.016$); \textsc{WebOverlay-OOD} ($N{=}592$) has zero overlap with the training set by construction.

\subsection{Stratified Aspect-Ratio Slice Analysis}
\label{subsec:slices}
To assess robustness to page height, Table~\ref{tab:slice_results} reports both slices at the same fixed threshold:
\begin{itemize}[leftmargin=*]
    \item \textbf{Test-Common ($N{=}1{,}873$)}: On standard pages, our model ($88.3\%$ F1) trails the teacher by $2.9$ points (paired $95\%$ CI $[-5.7, -0.03]$, just excluding zero), while MobileNetV4 stays near $34\%$.
    \item \textbf{Test-Hard ($N{=}491$)}: On long pages, our tiles keep recall above $90\%$ and achieve the highest ROC-AUC of the score-producing models ($96.9\%$). The teacher has a higher F1 on this slice ($75.0\%$ vs.\ $70.3\%$; CI of the gap $[-16.7, +7.5]$) owing to its higher precision, but misses significantly more overlays (recall gap $+20.9$ points, CI $[+7.3, +35.1]$). This ordering is measured against teacher-assisted labels and reverses under blind labels (\S\ref{subsec:blind}). The lightweight baselines fall below $30\%$ F1.
\end{itemize}

\begin{table}[t]
\centering
\caption{Stratified slice evaluation on WebOverlay-2.4K at the fixed threshold $\tau=0.5$. The MLLM emits binary verdicts, so its AUC is omitted.}
\label{tab:slice_results}
\resizebox{\columnwidth}{!}{%
\begin{tabular}{llccccc}
\toprule
\textbf{Slice} & \textbf{Model} & \textbf{P (\%)} & \textbf{R (\%)} & \textbf{F1 (\%)} & \textbf{Acc (\%)} & \textbf{AUC (\%)} \\
\midrule
\multirow{4}{*}{\textit{Test-Common} ($N=1,873$)}
& MobileNetV4 (Vision Only) & 21.55 & 81.17 & 34.06 & 36.73 & 60.87 \\
& YOLO-style ConvNet & 31.62 & 60.21 & 41.46 & 65.78 & 68.14 \\
& Commercial MLLM (Teacher) & 85.58 & \textbf{97.61} & \textbf{91.20} & \textbf{96.21} & --- \\
& \textbf{Ours (3-Tile, Max-OR)} & \textbf{90.30} & 86.47 & 88.35 & 95.41 & \textbf{95.18} \\
\midrule
\multirow{4}{*}{\textit{Test-Hard} ($N=491$)}
& MobileNetV4 (Vision Only) & 10.88 & 86.05 & 19.32 & 37.07 & 72.55 \\
& YOLO-style ConvNet & 18.18 & 60.47 & 27.96 & 72.71 & 74.61 \\
& Commercial MLLM (Teacher) & \textbf{81.08} & 69.77 & \textbf{75.00} & \textbf{95.93} & --- \\
& \textbf{Ours (3-Tile, Max-OR)} & 57.35 & \textbf{90.70} & 70.27 & 93.28 & \textbf{96.93} \\
\bottomrule
\end{tabular}%
}
\end{table}

\begin{table}[t]
\centering
\caption{Recall and precision (\%) by aspect-ratio bucket $\rho = H/W$ on WebOverlay-2.4K ($\tau=0.5$, teacher-assisted reference). \textit{1-view}: the same student scoring one whole-page view. The $\rho \ge 8$ bucket contains only 22 overlays.}
\label{tab:recall_rho}
\resizebox{\columnwidth}{!}{%
\begin{tabular}{lccccccc}
\toprule
 & & & \multicolumn{2}{c}{\textbf{Teacher}} & \multicolumn{2}{c}{\textbf{Ours, 3-tile}} & \textbf{Ours, 1-view} \\
\cmidrule(lr){4-5} \cmidrule(lr){6-7} \cmidrule(lr){8-8}
\textbf{Aspect Ratio} $\rho$ & \textbf{Pages} & \textbf{Overlays} & \textbf{R} & \textbf{P} & \textbf{R} & \textbf{P} & \textbf{R} \\
\midrule
$[0, 2)$ & 793 & 161 & \textbf{99.4} & 82.1 & 83.9 & 93.8 & 80.7 \\
$[2, 4)$ & 804 & 162 & \textbf{97.5} & 86.8 & 86.4 & 90.9 & 71.6 \\
$[4, 8)$ & 492 &  75 & 89.3 & 90.5 & \textbf{96.0} & 77.4 & 62.7 \\
$\ge 8$  & 275 &  22 & 59.1 & 81.2 & \textbf{81.8} & 47.4 & 36.4 \\
\bottomrule
\end{tabular}%
}
\end{table}

\noindent\textbf{Recall Across Aspect Ratios}: Table~\ref{tab:recall_rho} splits recall by $\rho$. Teacher recall is $99.4\%$ on $\rho < 2$, falls to $89.3\%$ on $\rho \in [4, 8)$, and to $59.1\%$ on $\rho \ge 8$. The 1-view student loses recall earlier and faster ($71.6\%$ on $\rho \in [2, 4)$, $36.4\%$ on $\rho \ge 8$). The 3-tile student stays between $81.8\%$ and $96.0\%$ in every bucket ($18/22$ overlays on $\rho \ge 8$ vs.\ $13/22$ for the teacher, albeit at $47.4\%$ precision on that small bucket). To verify that the teacher's recall drop stems from input resolution rather than model capacity, we also scored the teacher on the same three viewport crops (Max-OR): its recall on $\rho \ge 8$ rebounds from $59.1\%$ to $95.5\%$ ($21/22$ overlays, and $100\%$ recall / $84.8\%$ F1 on blind Test-Hard), at $3\times$ API cost ($\$1{,}143$/M pages).

\subsection{Teacher-Free Blind Evaluation}
\label{subsec:blind}
The comparisons above use teacher-assisted labels, which favor the teacher (\S\ref{subsec:setup}). To compare student and teacher without this bias, Table~\ref{tab:blind} scores both paradigms against \textsc{WebOverlay-OOD}, an independent blind-annotated benchmark ($N=592$) whose labels were assigned from full screenshots alone without access to any model predictions. We report paired bootstrap $95\%$ confidence intervals ($10{,}000$ resamples) for the F1 difference and an exact two-sided McNemar test on discordant pairs.

\begin{table}[t]
\centering
\caption{Student vs.\ teacher on the blind out-of-distribution benchmark \textsc{WebOverlay-OOD} ($\tau = 0.5$, $N=592$). Labels were assigned from full screenshots alone without access to any model predictions. $\Delta$: student minus teacher F1, with paired bootstrap $95\%$ CI ($10{,}000$ resamples). $p$: exact McNemar binomial test on discordant pairs.}
\label{tab:blind}
\resizebox{\columnwidth}{!}{%
\begin{tabular}{lcccccc}
\toprule
\textbf{Slice} & \textbf{$N$} & \textbf{Pos.} & \textbf{Teacher F1} & \textbf{Ours F1} & \textbf{$\Delta$ [95\% CI]} & \textbf{$p$} \\
\midrule
All & 592 & 133 & 84.4 & \textbf{87.9} & $+3.6$ [$-1.4$, $+8.6$] & 0.200 \\
Val-Common & 475 & 114 & \textbf{90.4} & 88.2 & $-2.3$ [$-6.6$, $+1.9$] & 0.541 \\
Val-Hard & 117 & 19 & 30.8 & \textbf{86.5} & $\mathbf{+55.7}$ [$+33.1$, $+80.4$] & $\mathbf{<0.001}$ \\
\bottomrule
\end{tabular}%
}
\end{table}

\begin{table*}[t]
\centering
\small
\caption{Ablation study on the human-verified \textbf{WebOverlay-2.4K} evaluation set ($N=2{,}364$) at the fixed operating point $\tau=0.5$, plus threshold-free AP and AUC. $\Delta$F1 is in absolute points relative to the reference row of each group. (A) Distillation depth at matched tiling; (B) tiling strategy; (C) cross-tile aggregation. Groups (B) and (C) apply the same trained weights with different inference-time inputs or aggregation rules; nothing is retrained. In (C), the F1 drop at $\tau = 0.5$ reflects calibration: AP and AUC are nearly unchanged.}
\label{tab:ablation_study}
\resizebox{\textwidth}{!}{%
\begin{tabular}{llccccccc}
\toprule
\textbf{Group} & \textbf{Model Variant / Ablated Component} & \textbf{Precision (\%)} & \textbf{Recall (\%)} & \textbf{F1 (\%)} & \textbf{Acc (\%)} & \textbf{AP (\%)} & \textbf{AUC (\%)} & $\mathbf{\Delta}$\textbf{F1 (pts)} \\
\midrule
\multirow{4}{*}{\textit{(A) Distillation Depth}}
& \textbf{Full 27-Layer Distillation (3-Tile, Ours)} & 85.08 & \textbf{86.90} & \textbf{85.98} & \textbf{94.97} & \textbf{87.78} & \textbf{94.95} & \textbf{Ref} \\
& Top-4 Layers Only (3-Tile) & 85.04 & 81.19 & 83.07 & 94.12 & 86.85 & 94.43 & $-2.91$ \\
\cmidrule{2-9}
& Full 27-Layer Distillation (Whole-Page View) & \textbf{94.65} & 71.67 & 81.57 & 94.25 & 85.91 & 91.32 & Ref \\
& Top-4 Layers Only (Whole-Page View) & 91.12 & 65.95 & 76.52 & 92.81 & 82.34 & 89.59 & $-5.05$ \\
\midrule
\multirow{5}{*}{\textit{(B) Tiling Strategy}}
& \textbf{3-Tile Viewport Tiling ($T_{\text{top}} + T_{\text{center}} + T_{\text{bottom}}$, Ours)} & 85.08 & \textbf{86.90} & \textbf{85.98} & \textbf{94.97} & \textbf{87.78} & \textbf{94.95} & \textbf{Ref} \\
& Whole-Page View (1 Tile) & \textbf{94.65} & 71.67 & 81.57 & 94.25 & 85.91 & 91.32 & $-4.41$ \\
& Bottom Tile Only ($T_{\text{bottom}}$) & 86.05 & 69.05 & 76.62 & 92.51 & 80.86 & 91.31 & $-9.37$ \\
& Center Tile Only ($T_{\text{center}}$) & 85.32 & 22.14 & 35.16 & 85.49 & 41.67 & 62.08 & $-50.82$ \\
& Top Tile Only ($T_{\text{top}}$) & 93.10 & 12.86 & 22.59 & 84.35 & 38.84 & 62.47 & $-63.39$ \\
\midrule
\multirow{3}{*}{\textit{(C) Cross-Tile Aggregation}}
& \textbf{Max-OR over Tile Scores (Ours)} & 85.08 & \textbf{86.90} & \textbf{85.98} & \textbf{94.97} & \textbf{87.78} & \textbf{94.95} & \textbf{Ref} \\
& Cross-Tile Feature Fusion (Mean + Max Features) & 92.19 & 14.05 & 24.38 & 84.52 & 86.71 & 94.57 & $-61.60$ \\
& Mean over Tile Scores & \textbf{94.55} & 12.38 & 21.89 & 84.31 & 85.79 & 94.58 & $-64.09$ \\
\bottomrule
\end{tabular}%
}
\end{table*}

On standard aspect-ratio pages (Val-Common), student and teacher are statistically indistinguishable ($p = 0.541$, CI spans zero). On the challenging slice (Val-Hard: $\rho > 8$, $H > 10{,}000$\,px, or non-English EU/JP ccTLDs), the commercial MLLM collapses to $30.8\%$ F1 (Recall $21.1\%$, missing 15 of 19 true overlays; $\kappa = 0.24$), mirroring the failure of the single-view vision model ($18.2\%$ F1, Recall $10.5\%$). In contrast, our 3-tile student achieves $86.5\%$ F1 at the fixed threshold $\tau=0.5$ ($88.9\%$ precision, $84.2\%$ recall, finding 16 of 19 overlays; $92.3\%$ F1 finding 18 of 19 at a calibrated threshold $\tau=0.05$), outperforming the teacher by $+55.7$ F1 points ($95\%$ CI $[+33.1, +80.4]$). An exact McNemar test on the 15 discordant pages at $\tau=0.5$ shows our model correct on 14 versus only 1 for the teacher ($p < 0.001$), consistent with the paired in-set blind Test-Hard result ($92.9\%$ vs.\ $63.6\%$ F1, discordant $6$--$0$, $p = 0.031$). This confirms that the teacher's recall degradation on long pages is severe, and demonstrates that rendering-anchored viewport tiling decisively overcomes the aspect-ratio bottleneck.

\begin{figure*}[t]
\centering
\includegraphics[width=\textwidth]{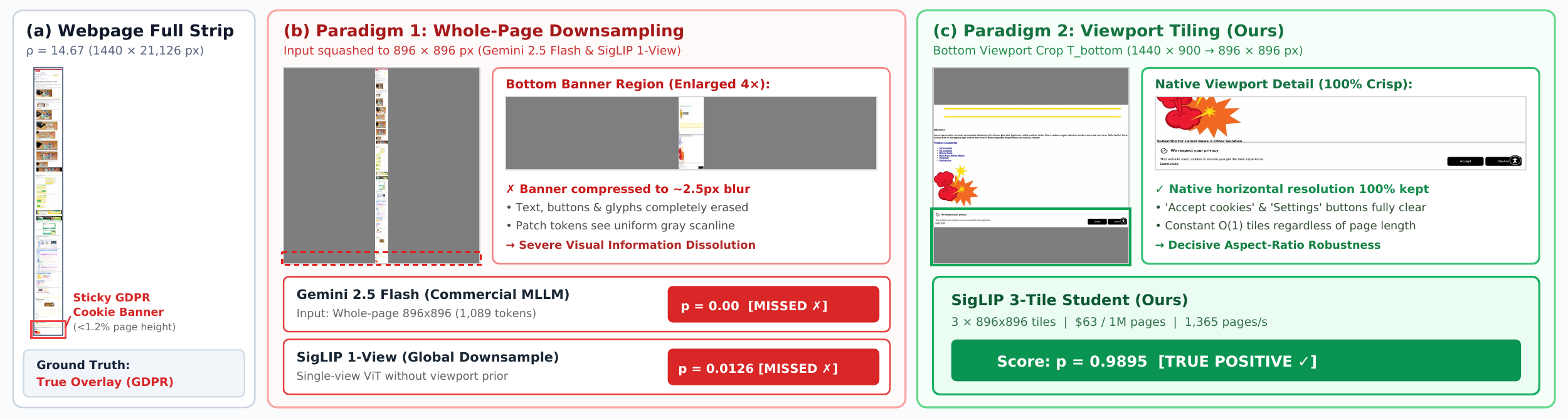}
\caption{\textbf{Qualitative comparison of extreme aspect-ratio collapse vs.\ rendering-anchored viewport tiling on an ultra-long web page} (\texttt{weboverlay\_val\_00196}, aspect ratio $\rho = 14.67$, $1440 \times 21{,}126$\,px). \textbf{Left}: Full-length webpage strip with ground-truth sticky GDPR cookie consent banner at the bottom ($<1.2\%$ of page height). \textbf{Middle}: Whole-page downsampling (longer side $1{,}280$\,px for Gemini 2.5 Flash, shrinking width to $87$\,px; $896 \times 896$ for 1-view SigLIP) compresses the banner into an illegible $\sim 2$\,px gray blur, causing both the commercial MLLM (binary verdict: clean) and 1-view vision model ($p=0.0126$) to fail with false negatives. \textbf{Right}: Our width-aligned bottom viewport tile $T_{\text{bottom}}$ ($1440 \times 1080 \to 896 \times 896$\,px) preserves native horizontal resolution and text legibility, enabling the 3-tile student network to confidently identify the overlay ($p=0.9895$, True Positive).}
\Description{Side-by-side visual comparison showing how whole page downsampling compresses bottom cookie banners to illegible pixels leading to MLLM false negatives, while viewport tiling preserves resolution and achieves confident detection.}
\label{fig:case_studies}
\end{figure*}

\subsection{Component Analysis \& Ablation Study}
\label{subsec:ablations}
Table~\ref{tab:ablation_study} isolates each design choice on the full evaluation set at the same fixed threshold, and adds threshold-free AP. Each setting is a single training run.

\noindent\textbf{Distillation Depth}: At matched tiling, restricting fine-tuning to the top four layers lowers F1 by $2.9$ points with three tiles ($83.1\%$ vs.\ $86.0\%$) and by $5.1$ points with the whole-page view, but lowers AP by only $0.9$ and $3.6$ points. We therefore read full adaptation as a modest, not decisive, benefit.

\noindent\textbf{Viewport Tiling vs.\ Whole-Page View}: Applied to a single whole-page view, the same weights reach high precision ($94.65\%$) but lose $15.24$ points of recall ($71.67\%$ vs.\ $86.90\%$; paired CI $[11.6, 19.0]$). The whole-page view still shows large translucent scrims, which keeps precision high, but downsampling long pages erases small dismiss controls (\S\ref{subsec:tiling}). The threshold-free gain is concentrated on long pages: AP rises by only $1.9$ points on the full set (CI $[-1.1, +4.8]$) but by $14.3$ points on Test-Hard (CI $[+0.3, +28.2]$), and ROC-AUC by $3.6$ and $14.7$ points. Among single tiles, the bottom tile alone reaches $76.6\%$ F1, whereas the top and center tiles alone reach $22.6\%$ and $35.2\%$; only their combination covers overlays at all three positions.

\noindent\textbf{Cross-Tile Aggregation}: Feature fusion and score averaging rank pages almost as well as Max-OR (AP $86.7\%$ and $85.8\%$ vs.\ $87.8\%$) but lose most of their recall at $\tau = 0.5$, because averaging dilutes the one tile that contains the overlay. Since these variants reuse the same weights without retraining, the large $\Delta$F1 in Table~\ref{tab:ablation_study}(C) reflects calibration at the fixed threshold rather than ranking quality. We choose Max-OR because it matches the any-tile semantics of overlays, keeps tiles independent for parallel execution, and attributes each decision to a specific tile, not because of this F1 gap.

\subsection{Qualitative Discrepancy \& Error Analysis}
\label{subsec:qualitative}
Figure~\ref{fig:case_studies} provides an empirical case study on an ultra-long web page from \textsc{WebOverlay-OOD} (\texttt{weboverlay\_val\_00196}, $\rho = 14.67$, $1440 \times 21{,}126$\,px), directly visualizing the physical failure mode of whole-page downsampling versus width-aligned viewport tiling. When the whole page is downscaled (longer side $1{,}280$\,px for Gemini 2.5 Flash, shrinking page width to $87$\,px, or $896 \times 896$ for 1-View SigLIP), the bottom GDPR cookie consent banner ($1440 \times 250$\,px) is crushed into an illegible $\sim 2$-pixel gray blur, causing Gemini 2.5 Flash to return a clean verdict (FN) and 1-View SigLIP to produce $p=0.0126$ (FN). In sharp contrast, our bottom viewport tile $T_{\text{bottom}}$ ($1440 \times 1080 \to 896 \times 896$\,px) preserves native horizontal resolution and text legibility, enabling the 3-tile student to detect the overlay with high confidence ($p=0.9895$, TP).

Beyond aspect-ratio dilution, two additional error sources recur across the benchmark: (1) \textbf{Semi-Transparent Scrims}, where full-screen dimmed backdrops lack localized box anchors, keeping ConvNet object detector recall near $60\%$; and (2) \textbf{Small Floating Widgets}, where benign chat bubbles trigger false alarms in HTML rules (\texttt{position: fixed}, \texttt{z-index}) and remain a residual false-positive source for vision models (\S\ref{subsec:limitations}).

\section{Industrial System Profiling \& Serving Economics}
\label{sec:deployment}

\subsection{Pipeline Latency Breakdown}
\label{subsec:profiling}
Table~\ref{tab:latency_breakdown} reports per-page timing of the 3-tile model on a single Cloud TPU v5e chip (bfloat16) with Intel Xeon host CPUs, measured under warm JAX JIT execution with batch size 16. The ViT encoder dominates; host-side decoding and tiling and the classification head are minor.

\begin{table}[t]
\centering
\caption{Per-page latency breakdown on TPU v5e (batch size 16, three $896^2$ tiles per page). ViT time is the batch forward time divided by the number of pages. Host-to-device transfer is excluded.}
\label{tab:latency_breakdown}
\resizebox{\columnwidth}{!}{%
\begin{tabular}{llcc}
\toprule
\textbf{Execution Stage} & \textbf{Hardware} & \textbf{Per Page (ms)} & \textbf{Share} \\
\midrule
Image decode + 3-tile tiling & Host CPU (16 threads) & 5.45 & 2.8\% \\
SigLIP ViT forward ($3 \times 896^2$) & TPU v5e & \textbf{190.04} & 97.2\% \\
MLP classification head & TPU v5e & 0.004 & $<0.01\%$ \\
\midrule
\textbf{Serial total} & & \textbf{195.49} & 100.0\% \\
\bottomrule
\end{tabular}%
}
\end{table}

\noindent\textbf{Key Takeaways}: CPU pre-processing can run in background threads overlapped with accelerator execution, and the classification head adds negligible overhead. Nearly all compute is spent in a single ViT encoder pass, with no autoregressive decoding.

\subsection{Throughput \& Cloud Economics}
\label{subsec:cost}
\noindent\textbf{Our Distilled Model}: Batching amortizes memory bandwidth and dispatch overhead. In stream mode (batch size 1) a page takes about $0.3$\,s; in batched mode (batch size 16) the amortized cost falls to about $0.19$\,s per page on one chip, or \textbf{\$63 per million pages} at the on-demand Cloud TPU v5e list price (\$1.20/chip-hour). Because pages are independent, throughput scales near-linearly across chips.

\noindent\textbf{Commercial MLLM Baseline}: Commercial multimodal APIs run a large autoregressive backbone behind a remote endpoint, with request latency ranging from hundreds of milliseconds to seconds depending on load and context. Billing scales with tokens. We price one $896^2$ image, an 81-token prompt, and a 12-token verdict per page (no thinking tokens) at standard, non-batch list prices as of October 2026, counting the $1{,}089$ image tokens we measured on a Gemini endpoint for Gemini models and the $765$ given by OpenAI's tile rule for GPT-4o. Our teacher, Gemini 2.5 Flash, then costs \$381 per million pages; Gemini 3.5 Flash-Lite, one of its listed replacements, has the same price. The cheapest current Gemini model, Gemini 3.1 Flash-Lite, costs \$311, and GPT-4o costs \$2{,}235.

\noindent Our distilled model is therefore $83\%$ cheaper than its teacher and about $80\%$ cheaper than the cheapest current Gemini model at list price; discounted tiers on either side (batch APIs, committed-use TPUs) are not modeled. It runs on self-managed accelerator hardware with predictable throughput, establishing the technical and economic feasibility of continuous crawl-scale overlay auditing.

\subsection{Two-Tier Cascade: A Measured Negative Result}
\label{subsec:cascade}
A natural way to reduce cost further is to front our model (Tier 2) with a cheap high-recall screen (Tier 1). We evaluated this directly by sweeping each candidate gate over its score range on all $2{,}364$ pages and measuring the fraction of traffic forwarded to Tier 2 and the fraction of human-labeled overlays that survive the gate. The result is largely negative, and we report it as an instructive failure mode for practitioners.

\noindent\textbf{MobileNetV4 cannot gate at all.} Because its scores collapse into a narrow band around $0.5$ (\S\ref{subsec:setup}), every threshold either forwards most traffic or discards many true overlays.

\noindent\textbf{The ConvNet detector gates, but with modest returns.} With a genuine score spread, the detector traces a usable frontier (Table~\ref{tab:cascade}). However, keeping gate recall high forwards almost all traffic, and halving Tier-2 traffic discards roughly a third of true overlays at Tier 1.

\begin{table}[t]
\centering
\small
\caption{Measured two-tier cascade (ConvNet detector as Tier 1, our model as Tier 2; $N{=}2{,}364$). \textit{Gate recall} upper-bounds end-to-end recall since Tier-1 rejections are unrecoverable.}
\label{tab:cascade}
\begin{tabular}{ccc}
\toprule
$\tau$ & \textbf{Traffic to Tier 2 (\%)} & \textbf{Gate Recall (\%)} \\
\midrule
--- & 100.0 & 100.0 \\
0.10 & 98.9 & 100.0 \\
0.20 & 88.6 & 95.0 \\
0.25 & 74.7 & 86.4 \\
0.30 & 62.6 & 79.0 \\
0.40 & 46.7 & 69.0 \\
0.50 & 36.4 & 60.2 \\
\bottomrule
\end{tabular}
\end{table}

\noindent\textbf{Takeaway.} Cascading pays only when the gate is both nearly free \emph{and} ranks well. Here the cheap gates rank poorly (Table~\ref{tab:threshold_free}), so any meaningful traffic reduction comes at an unrecoverable loss of recall. For audit-grade crawling, where a missed overlay silently corrupts downstream indexing and representation learning, we recommend deploying Tier 2 unconditionally, and treat two-tier cascading strictly as a cost-capped degraded mode for peak-load throttling.

\section{Conclusion and Discussion}
\label{sec:conclusion}

We presented a vision-only distillation framework for overlay detection on extreme aspect-ratio web captures: (1) full-document rendering pins viewport-anchored overlays to the top, center, or bottom, so three full-width tiles suffice at $\mathcal{O}(1)$ cost; (2) width-aligned cropping gives the student higher effective resolution on long pages than its MLLM teacher, matching it on standard pages and surpassing it on long pages ($p < 0.001$) under blind evaluation; and (3) pre-filling teacher verdicts induces measurable annotator anchoring. On \textbf{WebOverlay-2.4K}, our model reaches \textbf{86.0\% F1} and \textbf{94.9\% ROC-AUC} ($91\%$ recall on \textit{Test-Hard}) at \textbf{\$63/M pages} on TPU v5e ($83\%$ below teacher cost).

\subsection{Limitations and Future Work}
\label{subsec:limitations}
\label{subsec:future_work}
\begin{enumerate}[leftmargin=*]
    \item \textbf{User-Gesture-Gated Popups}: Our renderer captures timer- and scroll-triggered modals after network quiescence; exit-intent overlays require synthetic events.
    \item \textbf{Small Non-Intrusive Badges}: Tiny floating widgets (e.g., chat icons) do not block reading but are occasionally flagged due to fixed layering.
    \item \textbf{Labeling Modality \& Micro-Widgets}: While \textsc{WebOverlay-OOD} ($N=592$) confirms student superiority on long pages ($p < 0.001$), training supervision uses MLLM pseudo-labels. Inter-annotator consistency on micro-widgets warrants further categorization.
    \item \textbf{Mechanism Isolation \& Single-Run Reporting}: Tile positions are validated by activations rather than layout boxes, and bootstrap intervals capture evaluation sampling.
\end{enumerate}

\noindent\textbf{Future Work}: We identify three avenues: (1) \textit{tile-level supervision} to align resolution; (2) \textit{missing image detection}; and (3) \textit{popup removal} via vision-grounded DOM actions.

\renewcommand{\bibfont}{\footnotesize}
\bibliographystyle{ACM-Reference-Format}
\bibliography{references}

\end{document}